\RequirePackage{fix-cm}
\documentclass[twocolumn,10pt,final,natbib]{svjour3}          
\smartqed  
\usepackage{graphicx}
\usepackage{times}
\usepackage{epsfig}
\usepackage{graphicx}
\usepackage{amsmath}
\usepackage{amssymb}
\usepackage{multirow}
\usepackage{makecell}
\usepackage{tabularx}
\usepackage{dblfloatfix} 
\usepackage{booktabs}
\usepackage{siunitx}
\usepackage{multirow}

\usepackage[pagebackref=true,breaklinks=true,bookmarks=false,colorlinks=true,linkcolor=blue,citecolor=blue]{hyperref}

\def\makeheadbox{\relax}

\begin{document}

\title{Segmentation-Based Deep-Learning Approach for Surface-Defect Detection
}


\author{Domen Tabernik \and
        Samo \v{S}ela \and
        Jure Skvar\v{c} \and
        Danijel Sko\v{c}aj
}


\institute{D. Tabernik \and D. Sko\v{c}aj \at
              University of Ljubljana, Faculty of Computer and Information Science\\
              Ve\v{c}na pot 113, 1000 Ljubljana, Slovenia\\
              \email{domen.tabernik@fri.uni-lj.si, danijel.skocaj@fri.uni-lj.si} \\              
           \and
           S. \v{S}ela$^{1}$ \and J. Skvar\v{c}$^{2}$ \at
              $^{1}$Kolektor Group d. o. o., $^{2}$Kolektor Orodjarna d. o. o. \\
              Vojkova 10, 5280 Idrija, Slovenia\\              
}
\date{}



\maketitle

\begin{abstract}
Automated surface-anomaly detection using machine learning has become an interesting and promising area of research, with a very high and direct impact on the application domain of visual inspection. Deep-learning methods have become the most suitable approaches for this task. They allow the inspection system to learn to detect the surface anomaly by simply showing it a number of exemplar images. This paper presents a segmentation-based deep-learn\-ing architecture that is designed for the detection and segmentation of surface anomalies and is demonstrated on a specific domain of surface-crack detection. The design of the architecture enables the model to be trained using a small number of samples, which is an important requirement for practical applications. The proposed model is compared with the related deep-learn\-ing methods, including the state-of-the-art commercial software, showing that the proposed approach outperforms the related methods on the specific domain of surface-crack detection. The large number of experiments also shed light on the required precision of the annotation, the number of required training samples and on the required computational cost. Experiments are performed on a newly created dataset based on a real-world quality control case and demonstrates that the proposed approach is able to learn on a small number of defected surfaces, using only approximately 25-30 defective training samples, instead of hundreds or thousands, which is usually the case in deep-learning applications. This makes the deep-learning method practical for use in industry where the number of available defective samples is limited. The dataset is also made publicly available to encourage the development and evaluation of new methods for surface-defect detection.
\end{abstract}

\section*{Introduction}

\begin{figure*}
\centering
\includegraphics[width=0.90\textwidth]{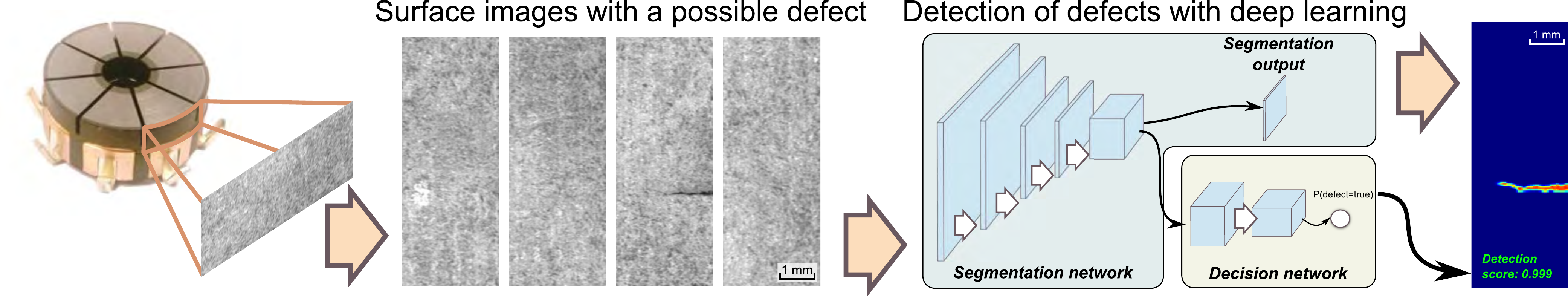}
\caption{The proposed scheme for the detection of surface defects\label{fig:intro}}
\end{figure*}

In industrial processes, one of the most important tasks when it comes to ensuring the proper quality of the finished product is inspection of the product's surfaces. Often, surface-quality control is carried out manually and workers are trained to identify complex surface defects. Such control is, however, very time consuming, inefficient, and can contribute to a serious limitation of the production capacity. 

In the past, classic machine-vision methods were sufficient to  address these issues~\citep{Paniagua2010,Bulnes2016}; however, with the Industry 4.0 paradigm the trend is moving towards the generalization of the production line, where rapid adaptation to a new product is required~\citep{Oztemel2018}. Classical machine-vision methods are unable to ensure such flexibility. Typically, in a classical machine-vision approach features must be hand-craft\-ed to suit the particular domain. A decision is then made using a hand-crafted rule-based approach or using learning-based classifiers such as SVM, decision trees or kNN. Since such classifiers are less powerful than deep-learning methods, the hand-craft\-ed features play a very important role. Various filter banks, histograms, wavelet transforms, morphological operations and other techniques are used to hand-craft appropriate features. Hand-engineering of features, therefore, plays an important role in classical approaches, but such features are not suited for different task and lead to long development cycles when machine-vision methods must be manually adapted to different products. A solution that allows for improved flexibility can be found in data-driven, machine-learning approaches where the developed methods can be quickly adapt\-ed to new types of products and surface defects using only the appropriate number of training images.   

This paper focuses on using state-of-the-art machine-learning methods to address the detection of visual surface defects. The focus is primarily on deep-learning methods that have, in recent years, become the most common approach in the field of computer vision. When applied to the problem of surface-quality control~\citep{Chen,FaghihRoohi2016,Weimer2013,Kuo2014}, deep-learning methods can achieve excellent results and can be adapted to different products. Compared to classical machine-vision methods, the deep learning can directly learn features from low-level data, and has higher capacity to represent complex structures, thus completely replacing hand-engineering of features with automated learning process. With a rapid adaptation to new products this method becomes very suitable for the flexible production lines required in Industry 4.0. Nevertheless, the open question remains: how much annotated data is required and how precise do the annotations need to be in order to achieve a performance suitable for practical applications? This is a particularly important question when dealing with deep-learning approaches as deep models with millions of learnable parameters often require thousands of images, which in practice is often difficult to obtain. 

This paper explores suitable deep-learning approaches for the surface-quality control. In particular, the paper studies deep-learning approaches applied to a surface-crack detection of an industrial product (see Fig.~\ref{fig:intro}). Suitable network architectures are explored, not only from their overall classification performance, but also from the point of view of three characteristics that are particularly important for Industry 4.0: (a) annotation requirements, (b) the number of required training samples and (c) computational requirements. The data requirement is addressed by utilizing an efficient approach with a deep convolutional network based on a two-stage architecture. Novel segmentation and decision network is proposed that is suited to learn from a small number of defected training samples, but can still achieve state-of-the-art results. 

An extensive evaluation of the proposed method is performed on a novel, real-world dataset termed Kolektor Sur\-face-Defect Dataset (KolektorSDD). The dataset represents a real-world problem of surface-defect detection for an industrial semi-finished product where the number of defective items available for the training is limited. The proposed approach is demonstrated to be already suitable for the studied application by highlighting three important aspects: (a) the required manual inspection to achieve a 100\% detection rate (by additional manual verification of detections), (b) the required details of annotation and the number of training samples leading to the required human labor costs and (c) the required computational cost. On the studied domain, the designed network is shown to outperform the related state-of-the-art methods, including the latest commercial product and two standard segmentation networks.

The remainder of the paper is structured as follows. The related work is presented in ``\nameref{sec:related-work}'' section, with details of the segmentation and decision net in~``\nameref{sec:method}'' section. An extensive evaluation of the proposed network is detailed in~``\nameref{sec:eval-seg}'' section, and a comparison with the state-of-the-art commercial solution is presented in~``\nameref{sec:eval-sota}'' section. The paper concludes with a discussion in ``\nameref{sec:conclusion}'' section.

\section*{Related work}\label{sec:related-work}

Deep-learning methods began being applied more often to surface-defect classification problems shortly after the introduction of AlexNet~\citep{Krizhevsky2012}. The work by~\cite{Masci2012} showed that for surface-defect classification the deep-learning approach can outperform classic machine-vi\-sion approaches where hand-engineered features are combined with support vector machines. They demonstrated this on the image classification of several steel defect types using a convolutional neural network with five layers. They achieved excellent results; however, their work was limited to a shallow network, as they did not use ReLU and batch normalization. A similar architecture was used by~\cite{FaghihRoohi2016} for the detection of rail-surface defects. They used ReLU for the activation function and evaluated several network sizes for the specific problem of classifying rail defects.

In a modern implementation of convolutional networks \cite{Chen} applied the OverFeat~\citep{Sermanet} network to detect five different types of surface errors. They identified a large number of labeled data, as an important problem for deep networks, and proposed to mitigate this using an existing pre-trained network. They utilized the OverFeat network trained on 1.2 million images of general visual objects from the ILSVRC2013 dataset and used it as feature extractor for the images with surface defects. They utilized a support vector machine to learn the classifier on top of deep features and showed that pre-trained features outperform LBP features. With the proposed Approximate Surface Roughness heuristic they were able to further improve on that result; however, their method does not learn the network on the target domain and is therefore not using the full potential of deep learning.

\cite{Weimer2016} evaluated several deep-learning architectures with varying depths of layers for surface-anomaly detection. They applied networks ranging from having only 5 layers to a network having 11 layers. Their evaluation focused on 6 different types of synthetic errors and showed that the deep network outperformed any classic method, with an average accuracy of 99.2\% on the synthetic dataset. Their approach was also able to localize the error within several pixels of accuracy; however, their approach to localization was inefficient as it extracted small patches from each image and classified each individual image patch separately. 

A more efficient network for explicitly performing the segmentation of defects was proposed by \cite{Racki2018}. They implemented a fully convolutional network with 10 layers, using both ReLU and batch normalization to perform the segmentation of the defects. Furthermore, they proposed an additional decision network on top of the features from the segmentation network to perform a per-image classification of a defect's presence. This allowed them to improve the classification accuracy on the dataset of synthetic surface defects.

Recently,~\cite{Lin2018} proposed the LEDNet architecture for the detection of defects on images of LED chips using a dataset with \num{30000} low-resolution images. Their proposed network follows the AlexNet architecture, but removes the fully connected layers and instead incorporates class-activation maps (CAMs), similar to~\citep{Zhou2015}. This design allows them to learn using only per-image labels and using CAMs for the localization of the defects. The proposed LEDNet showed a significant improvement in the defect-detection rate compared to traditional methods.  

Compared to the related methods, the approach proposed in this paper follows a two-stage design with the segmentation network and the decision network, similar to the architecture by~\cite{Racki2018}. 
However, the proposed approach incorporates several changes to the architecture of the segmentation and decision networks with the goal to increase the receptive field size and to increase the network's ability to capture small details. As opposed to some related works~\citep{Racki2018,Weimer2016}, the proposed network is applied to real-world examples instead of using synthetic ones. The used dataset in this study also consists of only a small number of defective training samples (i.e., 30 defective samples), instead of hundreds~\citep{Racki2018,Weimer2016} or thousands~\citep{Lin2018}. This makes some related architectures, such as LEDNet~\citep{Lin2018}, that use only per-image annotation and a large batch size, inappropriate for the task at hand. Since a small number of samples makes the choice of the network design more important, this paper evaluates the effect of replacing the segmentation network with two different standard network designs, normally used for the semantic segmentation, namely with DeepLabv3+~\citep{Chen2018} and U-Net~\citep{Ronneberger2015}. The impact of using pre-trained models is also evaluated by using the DeepLabv3+ network that is pre-trained on over 1.2 million images from the ImageNet~\citep{Russakovsky2015} and the MS COCO \citep{Lin2014} datasets.

\begin{figure*}
\includegraphics[width=\textwidth]{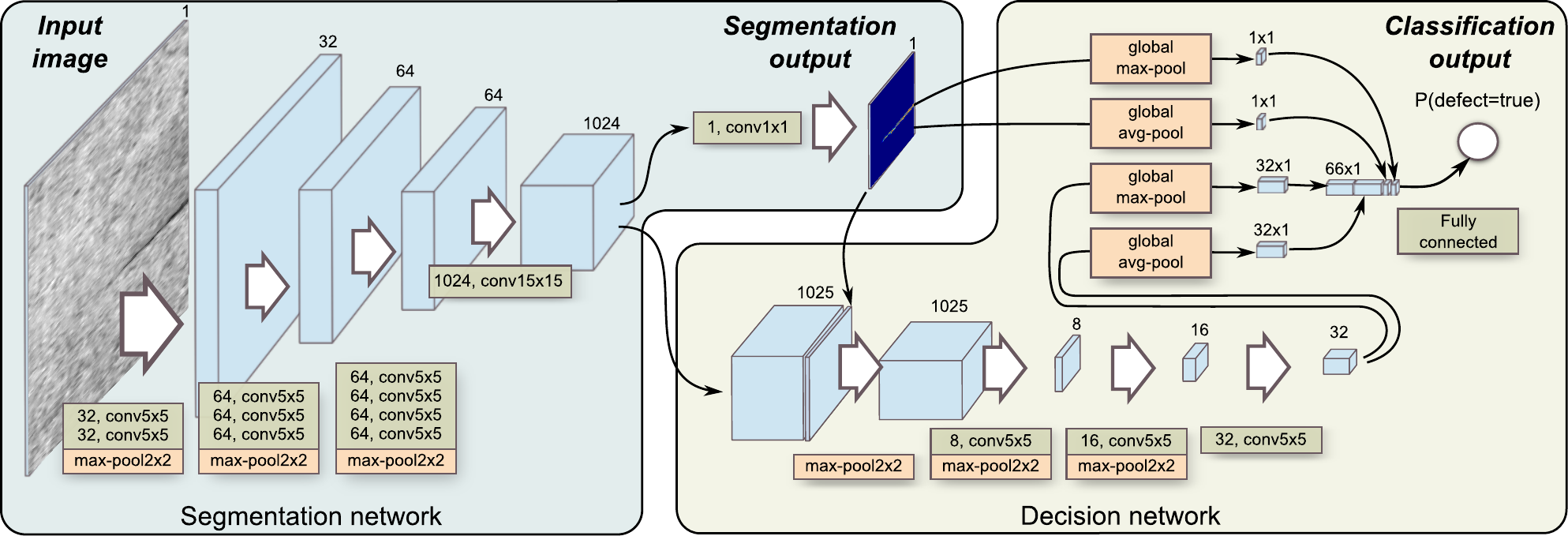}
\caption{The proposed architecture with the segmentation and decision networks\label{fig:arch}}
\end{figure*}
\section*{Proposed approach}\label{sec:method}

The problem of surface-anomaly detection is addressed as a binary-image-classification problem. This is suitable for surface-quality control, where an accurate per-image classification of the anomaly's presence is often more important than an accurate localization of the defect. However, to overcome the issue of a small number of samples in deep learning, the proposed approach is formulated as a two-stage design, as depicted in Fig.~\ref{fig:arch}. The first stage implements a segmentation network that performs a pixel-wise localization of the surface defect. Training this network with a pixel-wise loss effectively considers each pixel as an individual training sample, thus increasing the effective number of training samples and preventing overfitting. The second stage, where binary-image classification is performed, includes an additional network that is built on top of the segmentation network and uses both the segmentation output as well as features of the segmentation net. The first-stage network is referred to as \textit{the segmentation network}, while the second-stage network, as \textit{the decision network}. 

\subsection*{Segmentation network}

The proposed network consists of 11 convolutional layers and three max-pooling layers that each reduce the resolution by a factor of two. Each convolutional layer is followed by a feature normalization and a non-linear ReLU layer, which both help to increase the rate of convergence during the learning. Feature normalization normalizes each channel to a zero-mean distribution with a unit variance. The first nine convolutional layers use $5\times5$ kernel sizes, while the last two layers use $15\times15$ and $1\times1$ kernel sizes, respectively. A different number of channels is allocated for different layers, as can be seen in a detailed depiction of the network architecture in Fig.~\ref{fig:arch}. The final output mask is obtained after applying $1\times1$ convolution layer that reduces the number of output channels. This results in a single-channel output map with an 8-times-reduced resolution of the input image. Drop-out is not utilized in this approach, since the weight sharing in convolutional layers provides sufficient regularization.

The design of the proposed segmentation network focuses on the detection of small surface defects in a large-resolution image. To achieve this the network is designed with two important requirements: (a) the requirement for a large receptive field size in a high-resolution image and (b) the requirement to capture small feature details. This results in several significant changes of the architecture compared to the related work of \citep{Racki2018}. First, an additional down-sampling layer and large kernel sizes in higher layers are used to significantly increase the receptive field size. Second, the number of layers between each down-sampling is changed to having fewer layers in the lower sections of the architectures and having more layers in the higher sections. This increases the capacity of features with large receptive field sizes. Finally, the down-sampling is achieved using max-pooling instead of convolutions with a large stride. This ensures small but important details survive the down-sampling process, which is particularly important in this network with additional down-sampling layers.

\subsection*{Decision network}

The architecture of the decision network uses the output from the segmentation network as the input for the decision network. The network takes the output of the last convolutional layer of the segmentation network (1024 channels) concatenated with a single-channel segmentation output map. This results in 1025-channel volume that represents the input for the remaining layers with a max-pooling layer and a convolutional layer with $5\times5$ kernel sizes. Combination of both layers is repeated 3 times, with 8, 16 and 32 channels in the first, second and third convolutional layer, respectively. A detailed depiction of the architecture is given in Fig.~\ref{fig:arch}. The number of channels was chosen to increase as the resolution of the features decreases, therefore resulting in the same computational requirement for each layer. The proposed design effectively results in a 64-times-smaller resolution of the last convolutional layer than that of the original image. Finally, the network performs global maximum and average pooling, resulting in 64 output neurons. Additionally, the result of the global maximum and average pooling on the segmentation output map are concatenated as two output neurons, to provide a shortcut for cases where the segmentation map already ensures perfect detection. This design results in 66 output neurons that are combined with linear weights into the final output neuron.

The design of the decision network follows two important principles. First, the appropriate capacity for large complex shapes is ensured by using several layers of convolution and down-sampling. This enables the network to capture not only the local shapes, but also the global ones that span a large area of the image. Second, the decision network uses not only output feature volume of the last convolutional operation from the segmentation network before channel reduction with $1\times1$ kernel, but also the final segmentation output map obtained after the channel reduction with $1\times1$ kernel. This introduces a shortcut that the network can utilize to avoid using a large number of feature maps, if they are not needed. It also reduces the overfitting to a large number of parameters. The shortcuts are implemented at two levels: one at the beginning of the decision network where the segmentation output map is fed into several convolutional layers of the decision network, and another one at the end of the decision network where the global average and maximum values of the segmentation output map are appended to the input of the final fully-connected layer. The shortcut at the beginning of the decision network and the several convolutional layers with down-sampling are an important distinction with respect to the related work of \cite{Racki2018}. In contrast to the proposed work, they use only a single layer and no down-sampling in the decision layers, and do not use a segmentation output map directly in the convolution but only indirectly through global max and average pooling. This limits the complexity of the decision network and prevents it from capturing large global shapes.

\subsection*{Learning}

The \textit{segmentation network} is learned as a binary-seg\-men\-ta\-tion problem; therefore, the classification is performed at the level of individual image pixels. Two different training approaches were evaluated: (a) using a regression with a mean squared error loss (MSE) and (b) using a binary classification with a cross-entropy loss. The models are not pre-trained on other classification datasets, but instead are initialized randomly using a normal distribution.

The \textit{decision network} is trained with the cross-entropy loss function. Learning takes place separately from the segmentation network. First, only the segmentation network is independently trained, then the weights for the segmentation network are frozen and only the decision network layers are trained. By fine tuning only the decision layers the network avoids the issue of overfitting from the large number of weights in the segmentation network. This is more important during the stage of learning the decision layers than during the stage of learning the segmentation layers. The restrictions of the GPU memory limit the batch size to only one or two samples per batch when learning the decision layers, but when learning the segmentation layers each pixel of the image is considered as a separate training sample, therefore increasing the effective batch size by several folds.

\begin{figure*}
\centering
\includegraphics[width=\textwidth]{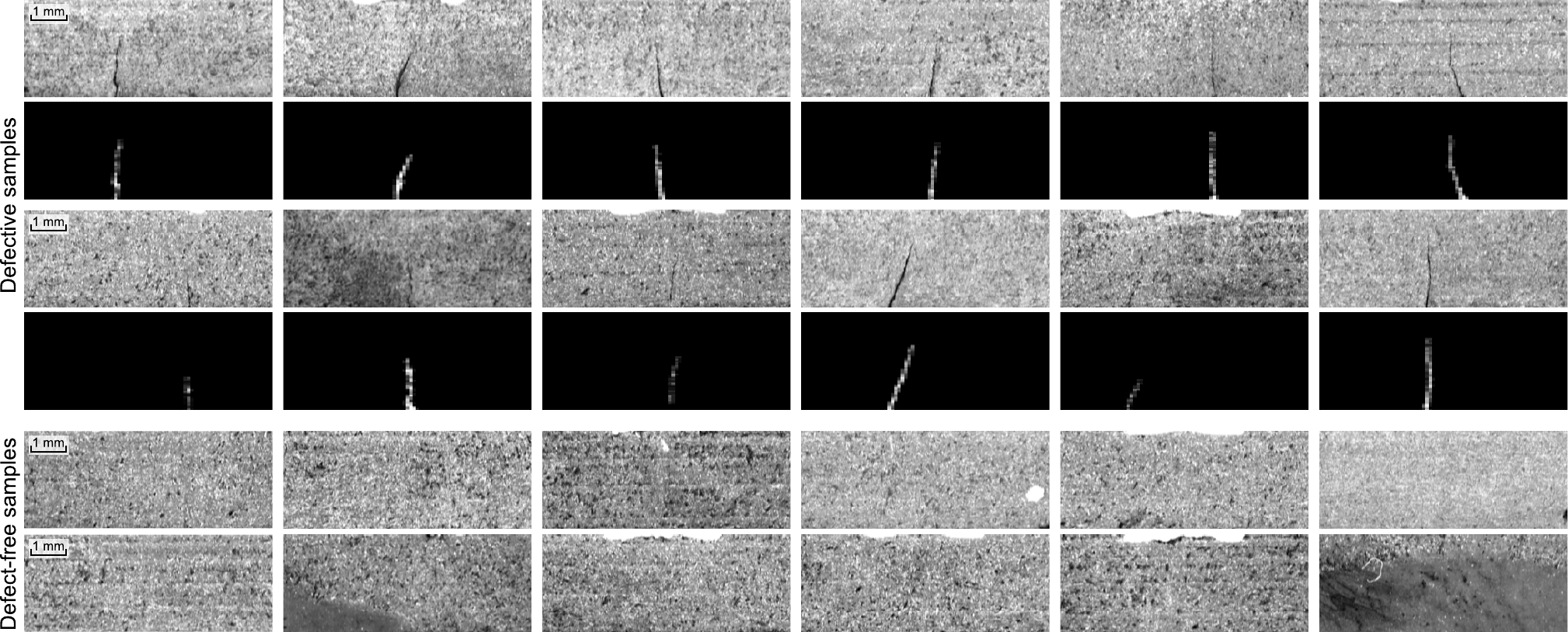}
\caption{Several examples of surface images with visible defects and their annotation masks in the top, and defect-free surfaces in the bottom\label{fig:db-examples}}
\end{figure*}

\begin{figure*}[!b]
\centering
\includegraphics[width=1\textwidth]{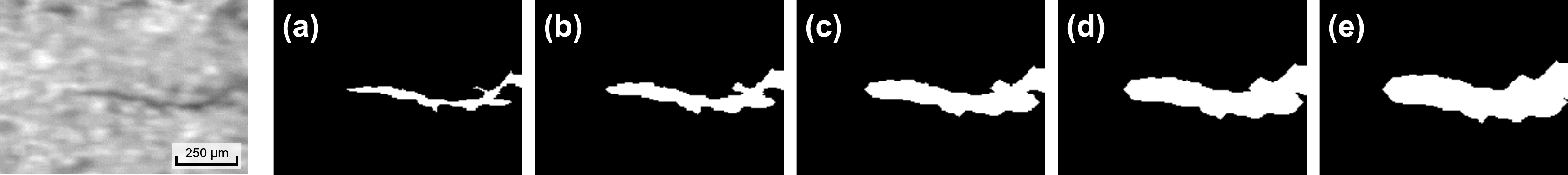}
\caption{\label{fig:annotations} Example of five different annotation types generated by dilating the original annotation shown in (a) with different morphological kernel sizes: (b) dilate=5, (c) dilate=9, (d) dilate=13 and (e) dilate=15}
\end{figure*}

The simultaneous learning of both the segmentation and decision networks was considered as well. The type of loss function played an important role in this case. Simultaneous learning was possible only when cross entropy was used for both networks. Since the losses are applied for different scopes, i.e., one at the per-pixel level and one at the per-image level, the accurate normalization of both layers played a crucial role. In the end, properly normalizing both losses proved not only more difficult to implement in practice than using a separate learning mechanism, but it also did not introduce any performance gain. The two-stage learning mechanism therefore proved to be a better choice and was subsequently employed in all experiments.

\subsection*{Inference} 

The input into the proposed network is a gray-scale image. The network architecture is independent of the input size, similar to fully convolutional networks~\citep{Long2015}, since fully connected layers are not used in feature maps, but only after the spatial dimension is eliminated with global average and max pooling. Input images can therefore be of a high or a low resolution, depending on the problem. Two image resolutions are explored in this paper: $1408\times512$ and $704\times256$.

The proposed network model returns two outputs. The first output is a segmentation mask as an output from the segmentation network. The segmentation mask outputs the probability of a defect for an $8\times8$ group of input pixels; therefore, the output resolution is reduced by 8 times with respect to the input resolution. The output map is not interpolated back to the original image size since the classification of $8\times8$ pixel blocks in high-resolution images suffices for the problem at hand. The second output is the probability score in the range of $[0,1]$ and represents the probability of an anomaly's presence in the image, as returned by the decision network.

\section*{Segmentation and decision network evaluation}\label{sec:eval-seg}

The proposed network is extensively evaluated on a surface-crack detection in an industrial product. This section first presents the details of the dataset and then presents the details of the evaluation and its results.

\subsection*{The Kolektor surface-defect dataset}

In the absence of publicly available datasets with real images of annotated surface defects a new dataset termed Kolektor surface-defect dataset (KolektorSDD) was created\footnote{The Kolektor surface-defect dataset is publicly available at http://www.vicos.si/Downloads/KolektorSDD}. The dataset is constructed from images of defected electrical commutators (see Fig.~\ref{fig:intro}) that were provided and annotated by Kolektor Group d. o. o.. Specifically, microscopic fractions or cracks were observed on the surface of the plastic embedding in electrical commutators. The surface area of each commutator was captured in eight non-overlapping images. The images were captured in a controlled environment, ensuring high-quality images with a resolution of $1408\times512$ pixels. The dataset consists of 50 defected electrical commutators, each with up to eight relevant surfaces. This resulted in a total of 399 images. In two items the defect is visible in two images while for remaining items the defect is only visible in a single image, which means there were 52 images where the defects are visible (i.e., defective or positive samples). For each image a detailed pixel-wise annotation mask is provided. The remaining 347 images serve as negative examples with non-defective surfaces. Examples of such images with visible defects and ones without them are depicted in Fig.~\ref{fig:db-examples}.

In addition, the dataset is annotated with several different types of annotations. This enables an evaluation of the proposed approach under different accuracies of the annotation. Annotation accuracy is particularly important in industrial settings since it is fairly time consuming and the human labor spent on annotation should be minimized. For this purpose, four more annotation types were generated by dilating the original annotations with the morphological operation using different kernel sizes, i.e., 5, 9, 13 and 17 pixels. Note that this is applied to images of the original resolution, and in the experiments with half the resolution the annotation mask is reduced after being dilated. All the annotations, one manual (a) and four generated ones (b-e), are depicted in Fig.~\ref{fig:annotations}.

\subsection*{Experiments}

The proposed network is first evaluated under several different training setups, which include different types of annotations, input-data rotation and different loss functions for the segmentation network. Altogether, the network was evaluated under four configuration groups:

\begin{itemize}
\item five annotation types,
\item two loss-function types for the segmentation network (mean squared error and cross entropy),
\item two sizes of input image (full size and half size),
\item without and with \ang{90} input-image rotation. 
\end{itemize}

\begin{figure*}
\sidecaption
\centering
\includegraphics[width=0.75\textwidth]{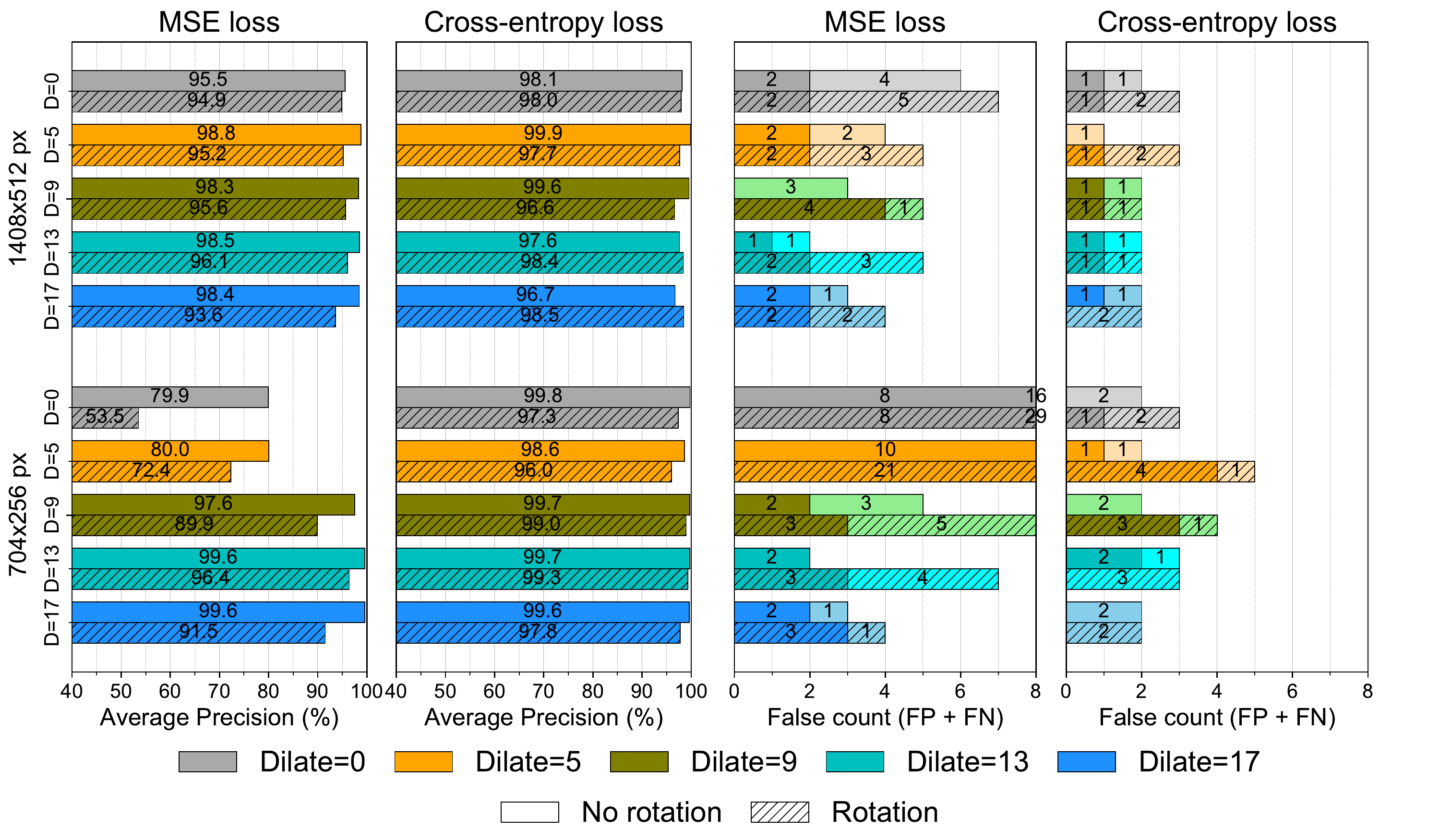}
\caption{Results of the proposed approach on the KolektorSDD (false positives (FP) shown in a dark colors and false negatives (FN) in light colors)\label{fig:decision-net-results}}
\end{figure*}

Each configuration group makes it possible to assess the performance of the network from four aspects. Different annotation types allow an assessment of the impact of the annotation's precision, while different image resolutions allow an assessment of the impacts on the classification performance at a lower computational cost. Additionally, the impact of different loss functions and the impact of augmenting the training data by rotating the images with the probability of 0.5 are also assessed.

For the purpose of this evaluation, the problem of surface-defect detection is translated into a binary-image-clas\-si\-fi\-ca\-tion problem. The main objective is to classify the image into two classes: (a) defect is present and (b) defect is not present. Although pixel-wise segmentation of the defect can be obtained from the segmentation network the evaluation does not measure the pixel-wise error, since it is not crucial in industrial settings. Instead, only the per-image binary-image-classification error is measured. The segmentation output is only used  for visualization purposes.

\subsubsection*{Performance metrics} The evaluation is performed with a 3-fold cross va\-li\-dation, while ensuring all the images of the same physical product are in the same fold and therefore never appear in the training and test set simultaneously. All the evaluated networks are compared considering three different classification metrics: (a) average precision (AP), (b) number of false negatives (FN) and (c) number of false positives (FP). Note, the positive sample is referred to as an image with a visible defect, and the negative sample, as an image with no visible defect. The primary metric used in the evaluation is average precision. This is more appropriate than FP or FN, since average precision is calculated as the area under the precision-recall curve and accurately captures the performance of the model under different threshold values in a single value. On the other hand, the number of miss-classifications (FP and FN) are dependent on the specific threshold applied to the classification score. We report the number of miss-cla\-ssi\-fi\-cations at a threshold value where the best F-measure is achieved. Also, note that AP was chosen instead of the area under the ROC curve (AUC) since AP more accurately captures the performance in datasets with a large number of negative (i.e., non-de\-fec\-tive) samples than does the AUC.

\subsubsection*{Implementation and learning details}

The network architecture was implemented in the TensorFlow framework \citep{tensorflow2015-whitepaper} and both networks are ~trained using a stochastic gradient descend without momentum. A learning rate of 0.005 was used for the mean squared error (MSE) and 0.1 for the cross-entropy loss. Only a single image per iteration was used, i.e., the batch size was set to one, mostly due to the large image sizes and the GPU memory limitations.

During the learning process the training samples were selected randomly; however, the selection process was modified to ensure that the network observed a balanced number of defective and non-defective images. This was achieved by taking images with defects for every even iteration, and images without defects for every odd iteration. This mechanism ensures that the system observes defective images at a constant rate; otherwise the learning is unbalanced in favor of the non-defective samples and would have learned significantly more slowly due to a larger set of non-defective images in the dataset. It should be noted that this leads to training that is not done exclusively by the epochs, as the number of non-defective images is 8-times higher than the number of defective ones and the network receives the same defective image before receiving all the non-defective images. 

Both networks were trained for up to 6600 steps. With 33 defective images per training set in one fold and alternating between defective and non-defective images in each step this translates to 100 epochs. One epoch is only considered to be over when  all the defective images are observed at least once, but not all the non-defective images are necessarily observed. 

\subsection*{Segmentation and decision network}

The proposed network that consists of both the segmentation network in the first stage and the decision network in the second stage is evaluated first. Detailed results are presented in Fig.~\ref{fig:decision-net-results}. This graph shows the results of experiments for different annotation types in different colors and experiments for using image rotation in dashed bars. The experiments for full image resolution are reported in the top group and experiments for half of the resolution at the bottom. The best-performing results were obtained with annotations dilated with $5\times5$ kernel sizes (\textit{dilate=5}), cross-entropy loss function, full image resolution and without any image rotations. The network in this configuration achieved an average precision (AP) of \num{99.9}{\%}, had zero false positive (FP) and one false negative (FN). 

\begin{figure}
\includegraphics[width=\columnwidth]{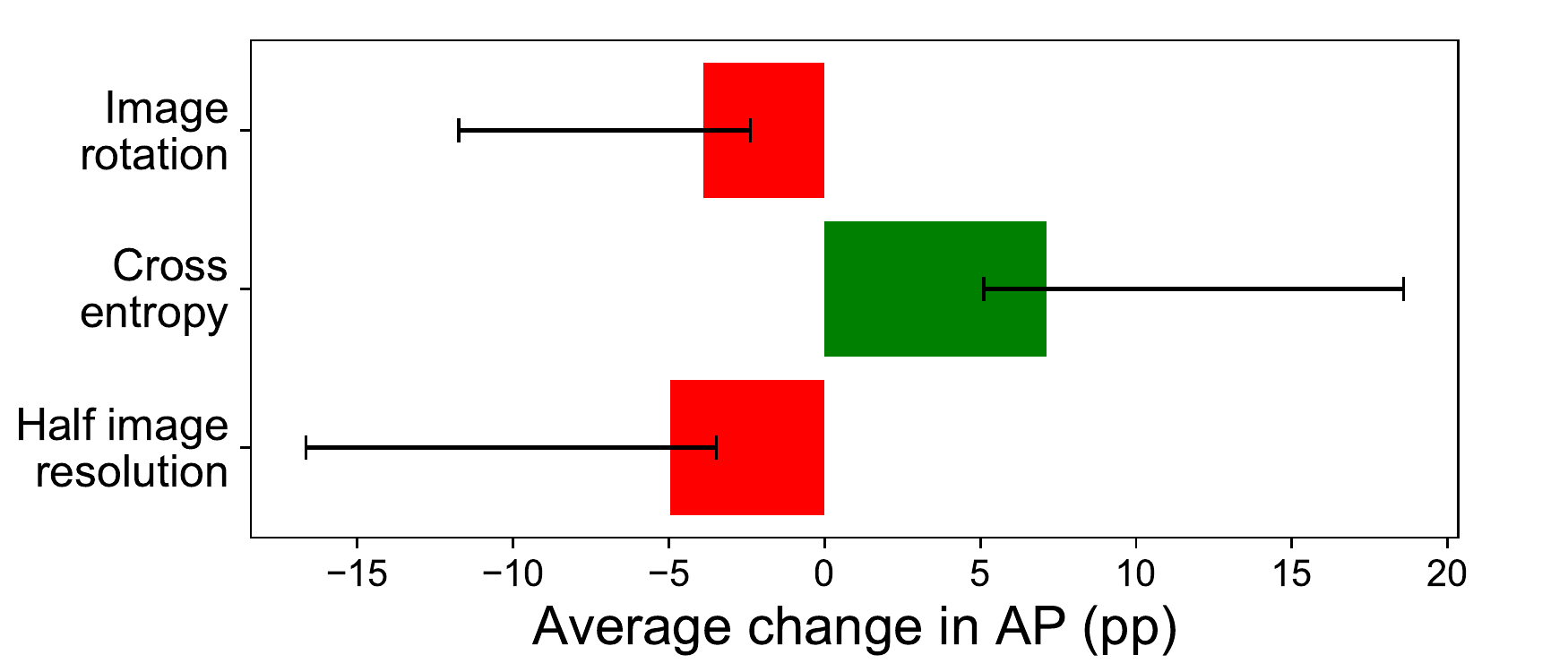}
\caption{Average changes in AP (average precision) as contributed by different changes to the learning configuration\label{fig:decision-net-settings-results}}
\end{figure}

Next, the impact of an individual learning setup can be assessed by observing the averaged improvement in performance for each specific change of the setting. An impact on the performance is reported for the following changes to the settings: (a) a change to the cross-entropy loss function for the segmentation network from a mean-squared-error loss, (b) a change to a smaller image resolution from the full image resolution, and (c) a change in the input data rotation by \ang{90} from no rotation. Improvements in AP averaged over all the experiments are reported in Fig.~\ref{fig:decision-net-settings-results}. The results for a specific change of setting, e.g., for a change to half the image resolution from the full image resolution, are obtained by first computing the AP of all the possible configurations of all the settings (reported in Fig.~\ref{fig:decision-net-results}) and then computing the differences in the AP between two experiments where only the setting in question was changed, e.g., between the experiment that used the half image resolution and one that used the full image resolution, but had all the other settings the same. The overall improvement in performance is captured through the average of the differences in AP over all the other settings that remained the same. Standard deviations are also reported separately  for the positive and negative directions.

\paragraph{Loss function}
When comparing the mean squared error loss (MSE) and the cross-entropy loss functions in Fig.~\ref{fig:decision-net-results} it is clear that the best performance is obtained with networks trained using the cross-entropy loss function. This is re\-flect\-ed in the AP metric and in the FP/FN count, as well as in improvements to the cross entropy averaged over all the other settings in Fig.~\ref{fig:decision-net-settings-results}. On average, the cross entropy achieved a 7-percent points (pp) better AP.

\begin{figure}
\includegraphics[width=\columnwidth]{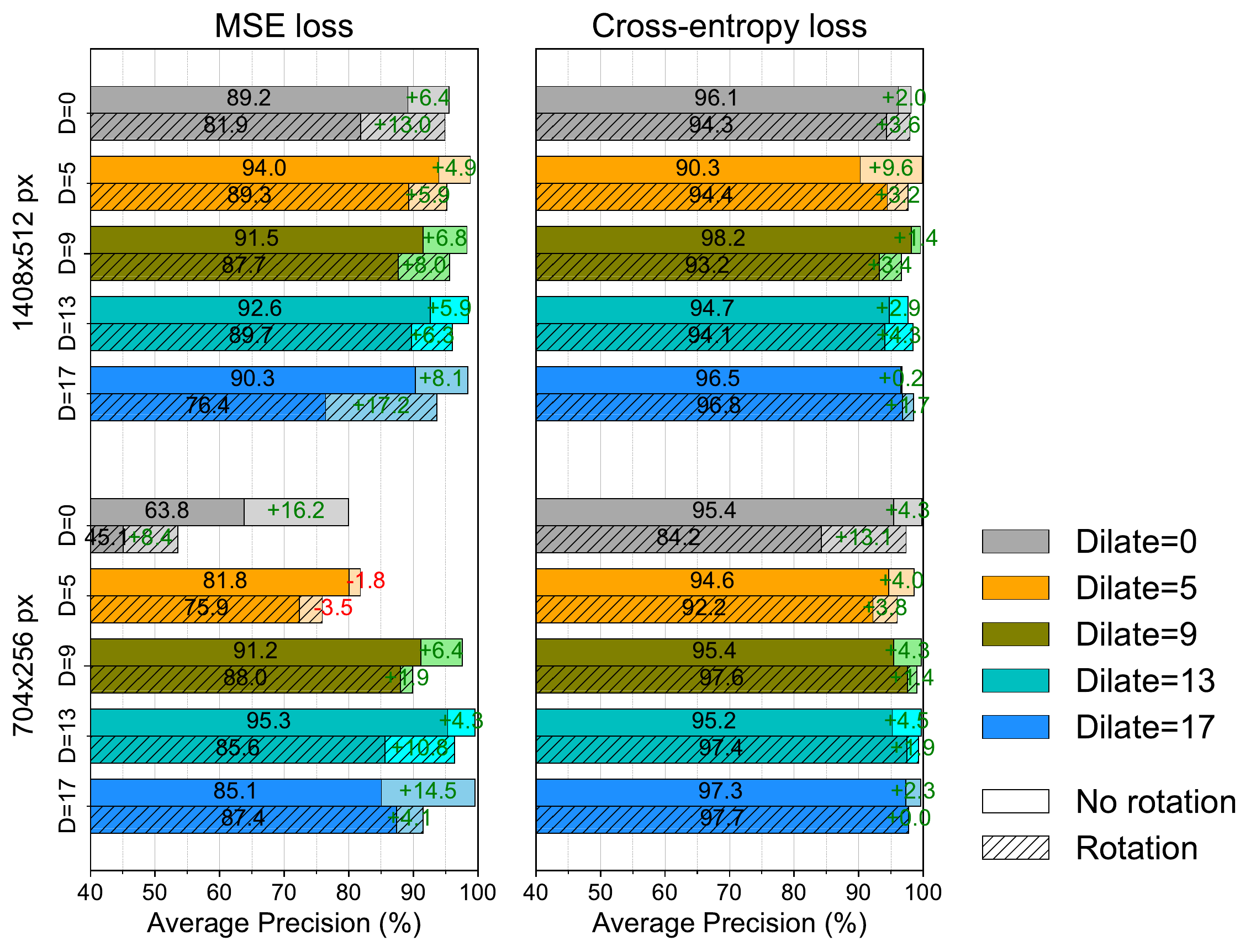}
\caption{Improvements in the average precision contributed by the decision network
\label{fig:seg-net-results}}
\end{figure}
\paragraph{Image resolution}
The network with the reduced image resolution on average performed with a 5-percent points worse AP as seen in Fig.~\ref{fig:decision-net-settings-results}. A close inspection of Fig.~\ref{fig:decision-net-results} shows that smaller images negatively impact mostly on networks trained with the MSE loss function, while networks trained with the cross entropy are not impacted. Cross entropy is less sensitive to the reduced image resolution and in some cases images with the reduced resolution perform marginally better (approximately one percent in AP). 

\paragraph{Image rotation}
Randomly rotating images, on the other hand, did not prove as useful and did not lead to any significant performance gains. In some cases the gain was at most one percent point; however, in other cases the performance was reduced by much more.

\paragraph{Annotation types} 
Finally, comparing different an\-no\-ta\-tion types in Fig.~\ref{fig:decision-net-results} results in only a slightly negative impact on the performance when training with smaller annotations (original or dilation with small kernels) and when considering the cross-entropy loss. The difference is more pronounced in the MSE loss function. Overall, the best results seem to be achieving annotations dilated with medium-to-large dilation rates.

\subsection*{Contribution of the decision network} 

The contribution of the decision network to the final performance is also evaluated. This contribution is measured by comparing the results from the previous section with the segmentation network without the decision network. Instead of the decision network, a simple two-dimensional descriptor and logistic regression are employed. A two-dimensional descriptor was created from the values of the global max and average pooling of the segmentation output map, which is then used as a feature for the logistic regression, which is learned separately from the segmentation network after the network has already been trained.

The results are presented in Fig.~\ref{fig:seg-net-results}. When focusing on models with a cross-entropy loss it is clear that the network with only the segmentation network already achieves fairly good results. The best configuration as obtained by \textit{dilate=9} annotation achieves an average precision (AP) of \num{98.2}{\%}, zero false positives (FP) and four false negatives (FN). The decision network, however, improves this result across most of the experiments. The contribution of the decision network is larger for the MSE loss. The average precision with the MSE loss function achieves an AP of less than 90\% when only the segmentation network is used, while with the decision network the AP is above 95\% for the MSE loss. For the network trained with the cross entropy the decision network contributes to the performance gain as well, but since the segmentation network already performs well, the improvements are slightly smaller, improving the AP by 3.6-percent points to more than 98\% on average for the decision network. The same trend is observed in the number of miss-classifications at the ideal threshold, where on average 4 miss-classifications for the segmentation network are reduced to 2 miss-classifications on average when the decision network is included.
 \begin{figure}
 \centering
 \includegraphics[width=\columnwidth]{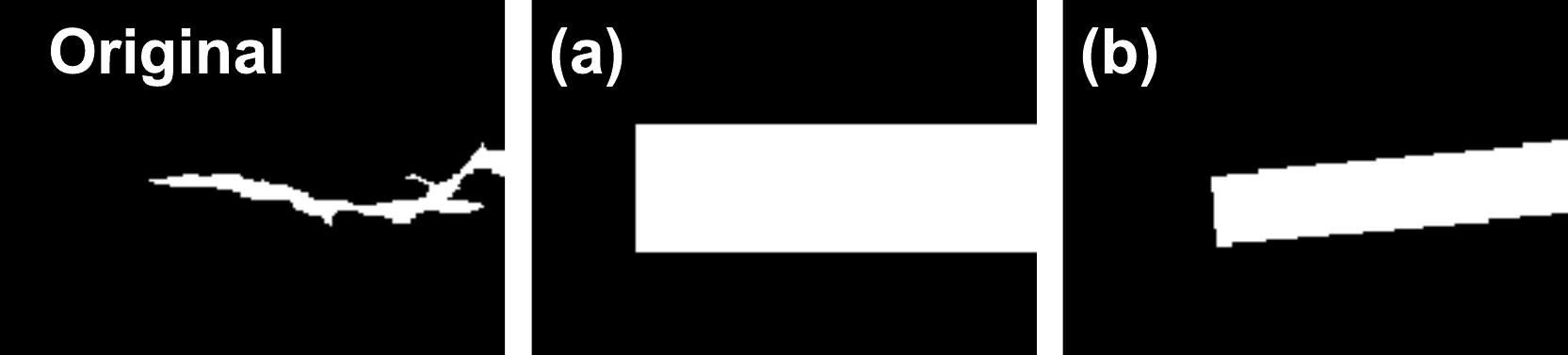}
 \caption{Two additional annotations: (a) big and (b) coarse \label{fig:big-coarse-annot}}
 \end{figure}

These results point to the important role of the decision network. Simple per-pixel output segmentation does not appear to have enough information to predict the presence of the defect in the image equally well as can the decision network. On the other hand, the proposed decision network is able to capture information from the rich features of the last segmentation layers, and through additional decision layers, it is able to separate the noise from the correct features. Additional down-sampling in the decision network have also contributed to the improved performance since this increased the receptive field size and enabled the decision network to capture the global shape of the defect. Global shape is important for the classification, but it is not important for the pixel-wise segmentation.

\subsection*{Required precision of the annotation}

Experiments from the previous section already demon\-strat\-ed that large annotations perform better than the finer ones. This is further explored in this section by assessing the impact of even coarser annotations on the classification performance. For this purpose two additional types of annotation were created, termed: (a) big annotation with a bounding box and (b) coarse annotation with a rotated bounding box. Both annotations are shown in Fig.~\ref{fig:big-coarse-annot}. This type of annotation is less time consuming for a human annotator to perform and would be better in an industrial setting. 

\begin{figure}
\centering
\includegraphics[width=0.95\columnwidth]{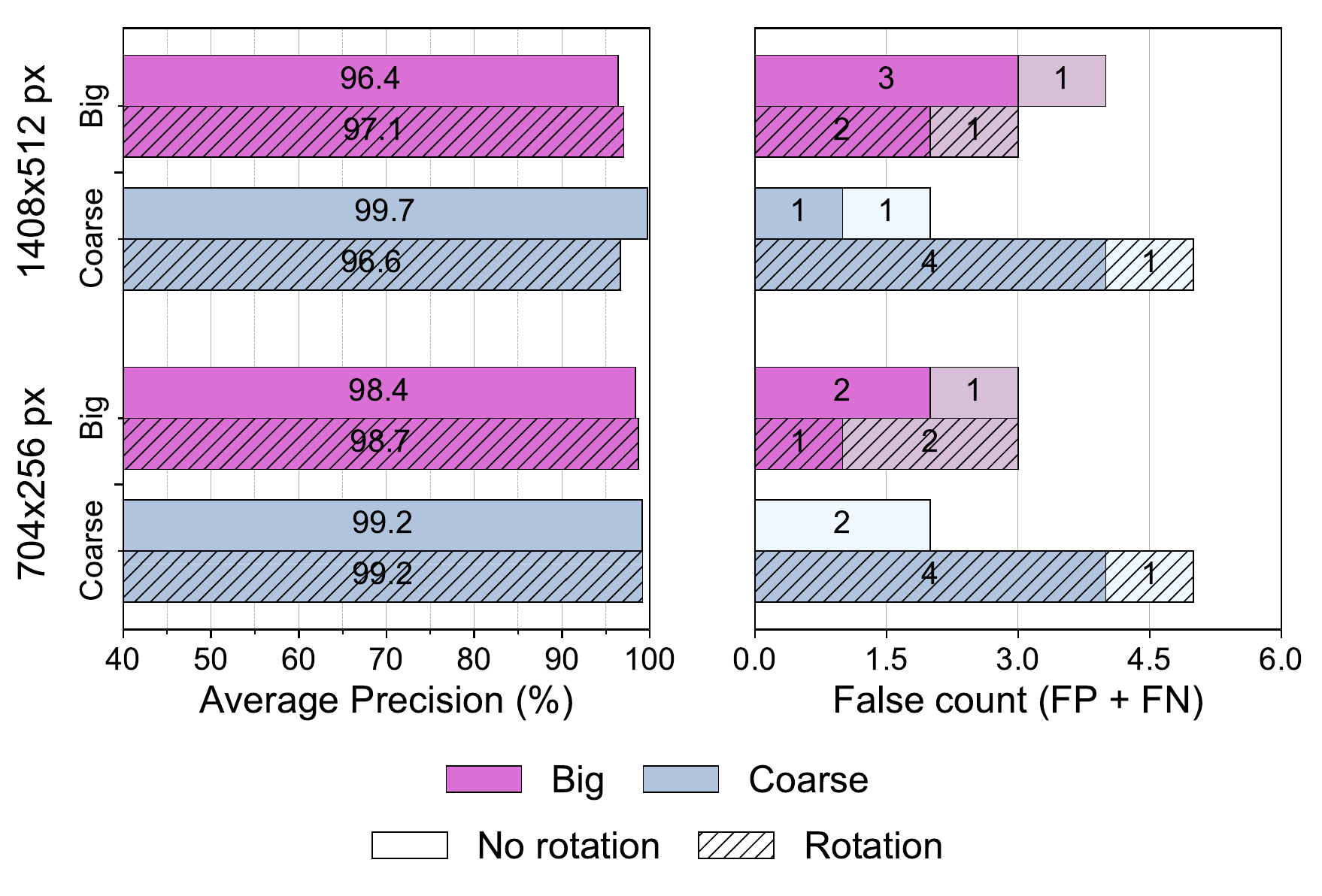}
\caption{Results with the \textit{big} and the \textit{coarse} annotations\label{fig:large-annot-results}}
\end{figure}

The results are presented in Fig.~\ref{fig:large-annot-results}. Only networks with a cross-entropy loss were used for this experiment, as the MSE loss proved less capable in previous experiments. The experiments show large annotations perform almost as well as the finer ones. The annotation denoted as \textit{big} performs slightly worse, with a best AP of 98.7\% and 3 miss-clas\-si\-fi\-ca\-tions, while the \textit{coarse} annotation achieves an AP of 99.7\% and 2 miss-clas\-si\-fi\-ca\-tions. Note that with a smaller image resolution both annotations achieve similar APs with the same number of miss-clas\-si\-fi\-ca\-tions. 

These results are comparable to the results obtained with finer annotations in the previous section where an AP of 99.9\% is achieved with only one miss-classification. Finer annotations do achieve slightly better results; however, considering that this level of detail is time consuming to annotate, it would still be feasible to use coarse annotations with minimal or no performance loss.

\begin{figure*}
\sidecaption
\includegraphics[width=0.70\textwidth]{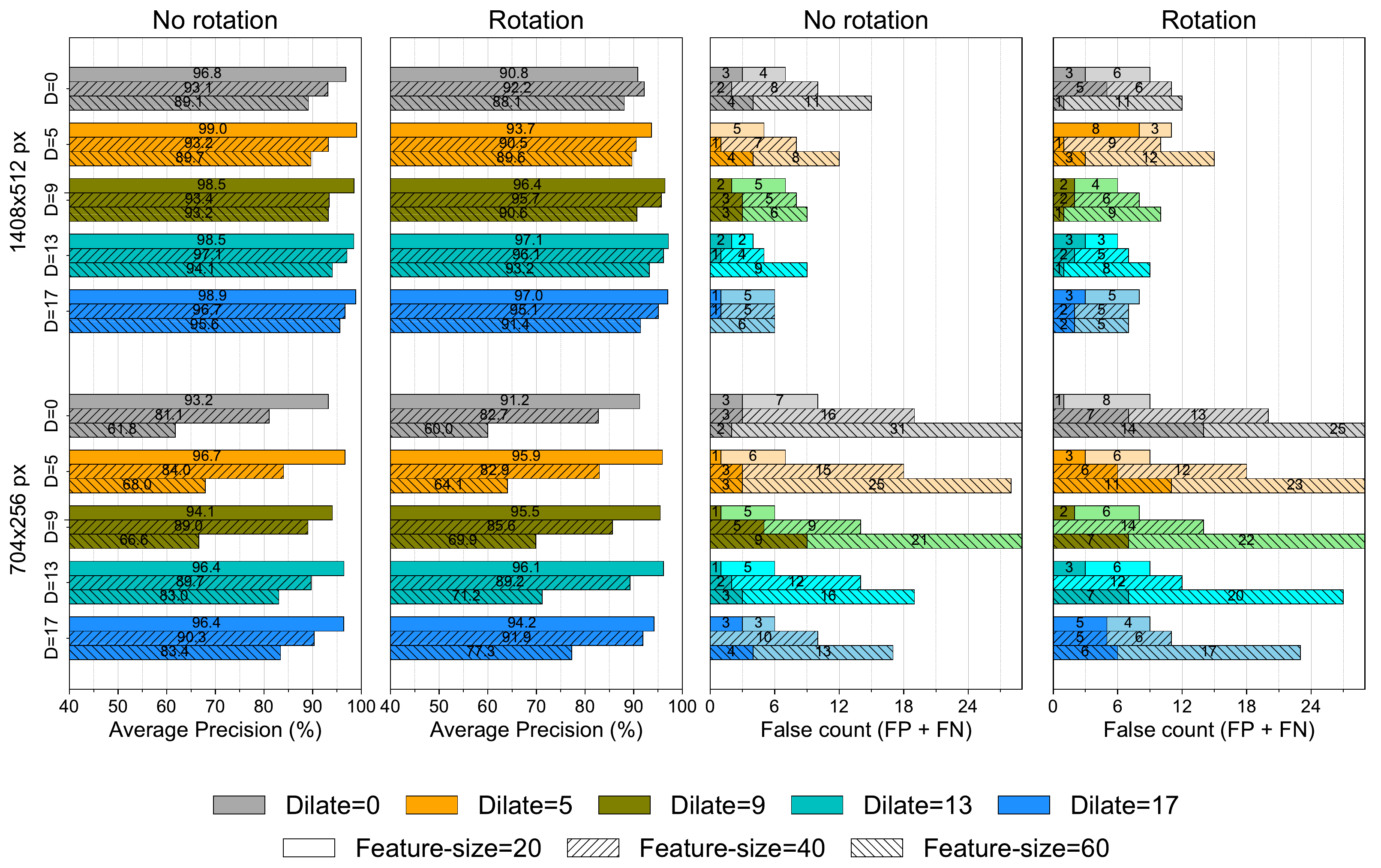}
\caption{Evaluation of the commercial software Cognex ViDi Suite on KolektorSDD\label{fig:vidi-results}}
\end{figure*}

\section*{Comparison with the state of the art}\label{sec:eval-sota}

Several state-of-the-art models are further evaluated to assess the performance of the proposed approach in the context of the related work. This section first demonstrates the performance of a state-of-the-art commercial product and two standard segmentation networks under different training configuration. This provides the best training configuration for each state-of-the-art method and allows for a fair comparison with the proposed network architecture, which is performed at the end of this section.

\subsection*{Commercial software}

The deep-learning-based state-of-the-art commercial software for industrial image analysis, Cognex ViDi Suite \citep{CognexVIDI}, is evaluated first. The Vidi company emerged from CSEM in 2012, a private, non-profit, Swiss research and technology organization, and was acquired by Cognex in 2017. The software package has three different deep-learning tools: ViDi blue (fixturing, localization), Vidi Red (segmentation and anomaly detection), ViDi green (object and scene classification). 

Vidi Red is a tool for anomaly detection, aesthetic visual inspection and segmentation. The tool can run in an unsupervised or supervised mode. In the former case only the images of non-defective samples are required, and in the latter only images of defective samples. The user can adjust various parameters from four distinctive regions: sampling (feature size, color), training (count epochs, train selection), perturbation (rotation, scale, aspect-ratio, shear, flip, luminance, contrast) and processing (sampling density, simple regions).

In this paper all the experiments using the Cognex ViDi Suite v2.1 were performed using the ViDi Red Tool in a supervised mode. The software is extensively evaluated under different learning configurations to find the best conditions for a fair comparison with the proposed approach in the next section. The following learning configurations are varied for this evaluation: 

%
%
%

\begin{itemize}
\item five annotation types,
\item three feature sizes (20, 40, 60 pixels),
\item two sizes of input image (full size and half size),
\item with/without \ang{90} input data rotation.
\end{itemize}

The settings that are varied are similar to ones in the ``\nameref{sec:eval-seg}'' section, with the difference being that different loss functions are not evaluated since the software does not provide such a detailed level of control. Instead, different sizes of the features are evaluated, which have proven to play a crucial role on the proposed dataset. Features of size from 20 to 60 pixels are evaluated. Features of less than 15 pixels are not recommended based on the documentation, while features larger than 60 pixels produced worse results.

\paragraph{Implementation details}
Access to the learning and inference core of the ViDi Suite is possible through the production and training API. All of the experiments were done in the C\#.Net programming language. The evaluation was performed with a 3-fold cross validation and the same train/test split as in the previous experiments, using a gray-scale image (number of color channels set to one) and learning for 100 epochs. The training was performed on all the images from the train fold; therefore, using a parameter \textit{training selection} of 100\%. The models were exported and evaluated in the production mode on the test folds with the parameter \textit{simple regions} enabled and the \textit{sampling density} set to one. This ensures an equivalent processing procedure as used in the proposed deep-learning model. We used default values as recommended by the vendor for the \textit{sampling density}. Experiments with values that represent denser sampling at the expense of slower inference were also performed, but this did not improve the result.

\paragraph{Results}

The results are presented in Fig.~\ref{fig:vidi-results}. Among the different learning setups, the best performance was achieved using the model trained with the \textit{dilate=5} annotations, using the smallest feature size (20 pixels), without rotating the images and using the original image size. The model achieved AP of \num{99.0}{\%}, and 5 miss-classifications, i.e., five FN and zero FP. Note that one model achieved only 4 miss-classifications, although with an overall lower AP.

\paragraph{Annotation sizes}
Among the different annotation types, the dilated annotations perform better than non-dilated ones. However, among the different dilation rates the performance gain is minimal with only \num{0.1}{pp} difference between \textit{dilate=5} and \textit{dilate=17}.

\begin{table*}[b]
\caption{Learning hyper-parameters with fixed learning values in the first three columns and the best selected learning configuration setup in the remaining four columns\label{tab:hyper-params}}
\centering
\begin{tabular}{lccccccccc}
\toprule
\textbf{\textit{Method}} & \makecell[c]{Number\\epochs} & \makecell[c]{Learning\\rate} & \makecell[c]{Initialization} & \makecell[c]{Annotation\\type} & \makecell[c]{Image size} & \makecell[c]{Data\\rot.} & \makecell[c]{Loss\\function} & \makecell[c]{Feature\\size} \\
\midrule
Segmentation/decision net (our) 		& \multirow{1}{*}{100} & \multirow{1}{*}{0.1} & \multirow{1}{*}{$\mathcal{N}(0.01)$} & Dilate=5 & \multirow{1}{*}{$1408\times512$} & \multirow{1}{*}{No} & \multirow{1}{*}{Cross-entropy} & \multirow{1}{*}{N/A} \\

U-Net~\citep{Ronneberger2015}	& 100 & 0.1 & $\mathcal{N}(0.01)$ & \multirow{1}{*}{Dilate=9} & $1408\times512$ & No & Cross-entropy& N/A \\
DeepLab v3+~\citep{Chen2018}	& 100 & 0.1 & \makecell[c]{Pre-trained} & Dilate=9  & $1408\times512$ & No & Cross-entropy & N/A \\
Cognex ViDi Suite 				& 100 & N/A & N/A & Dilate=5 & $1408\times512$ & No & N/A & 60 \\

\bottomrule
\end{tabular}
\end{table*}

\paragraph{Feature sizes}
Comparing the different feature sizes, the mod\-el with small features consistently outperforms models with larger features, regardless of the annotation precision. This can be contributed to the specifics of the dataset with a high image resolution and many small surface defects. Furthermore, experiments with the half-resolution image reveal that large features perform significantly worse than the smaller features in this case. This leads to the conclusion that large feature sizes cannot capture smaller details, which are important for the classification.

\paragraph{Image size and rotation}
Finally, the experiments also reveal that neither the half-resolution image nor randomly rotating the input data by \ang{90} results in an improved performance. The performance decreases slightly  in both cases, although the performance drop for both is minor.

\subsection*{Using state-of-the-art segmentation networks}\label{sec:eval-comercial}

Next, two standard segmentation networks are evaluated, namely, DeepLabv3+ \citep{Chen2018} and U-Net \citep{Ronneberger2015}. The DeepLab architecture was selected as a representative of the pre-trained model that achieves state-of-the-art results on current semantic seg\-men\-ta\-tion bench\-marks, while the U-Net architecture was selected as a representative of the models designed for a precise pixel-wise segmentation. The reader is referred to~\citep{Chen2018} for more detailed information about the DeepLabv3+ method and to~\citep{Ronneberger2015} for details about the U-Net model. Both models were evaluated under different annotations, but only cross entropy is considered for the loss function and only full-resolution image sizes without data rotation are used, since those settings proved to be the best performing in the previous experiments.

\paragraph{Implementation details}

Both segmentation methods were embedded into the proposed approach by replacing the segmentation part of the proposed network. A TensorFlow implementation of both networks was embedded in the proposed network. The DeepLabv3+ used in these experiments was based on the Xception~\citep{Chollet2017} architecture containing 65 convolutional layers, trained and evaluated on a single scale and using an output stride of 16. The U-Net used in these experiments was a modified U-Net architecture with 24 convolutional layers, where the only modification is an added batch normalization for each convolution. The original U-Net also outputs the segmentation in the full input resolution; however, since the pixel-wise accurate segmentation in full resolution is not in the interests of this experiment the output-map resolution was reduced by 8 times. This corresponds to the same output resolutions as in the proposed network. 

\begin{figure}
\centering
\includegraphics[width=1\columnwidth]{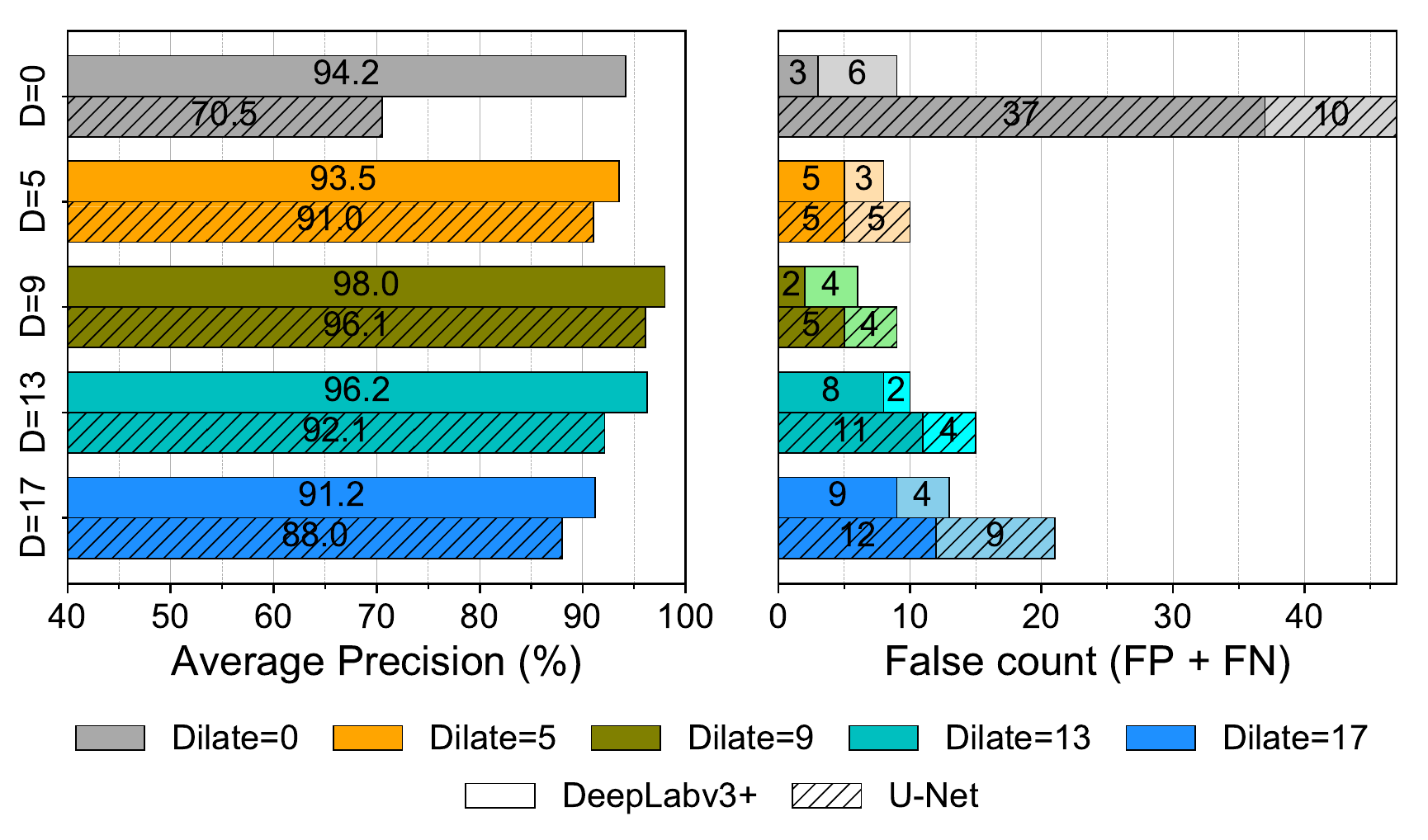}
\caption{Evaluation of two standard segmentation networks (DeepLabv3+ and U-Net)\label{fig:std-seg-results}}
\end{figure}

\begin{figure*}
\sidecaption
\centering
\includegraphics[width=0.80\linewidth]{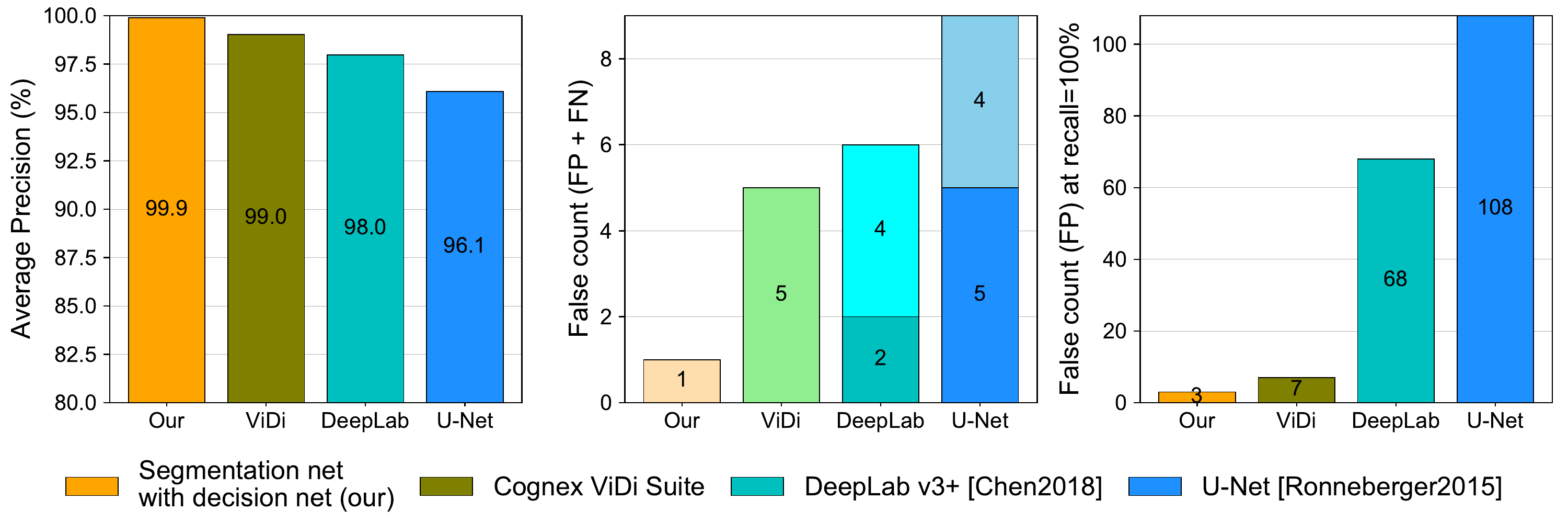}
\caption{Comparison with the state-of-the-art on KolektorSDD (in the middle graph: false positives in dark colors and false negatives in light)\label{fig:sota-results}}
\end{figure*}

For both segmentation networks the segmentation layers were trained separately from the decision layers, similar to the proposed approach, using 3-fold cross validation with the same train/test split as in all the previous experiments. Both methods were also evaluated with logistic regression that replaces the decision network, but this proved to perform worse. The parameters of the DeepLabv3+ network were initialized with a model that was pre-trained on the ImageNet~\citep{Russakovsky2015} and the COCO dataset \citep{Lin2014}, while the parameters of the U-Net network were initialized randomly using a normal distribution, similar to the initialization of the network presented in this paper. Both networks were trained for 100 epochs with the same learning procedure as used for the proposed model, i.e, using a learning rate of \num{0.1} without momentum, a batch size of 1, and alternating between defective and non-defective images for each step.

\paragraph{Results}
The results are shown in Fig.~\ref{fig:std-seg-results}. Of the standard networks, the best-performing model, i.e., DeepLabv3+ trained using \textit{dilate=9} annotations, achieved an AP of \num{98.0}{\%}, and obtained two FP and four FN at an ideal F-measure. Overall, slightly dilated annotations were shown to achieve the best results, while the annotations dilated with larger kernels gave worse results. On average, DeepLabv3+ also outperformed U-Net architecture by $2-3$ percent points in average precision, regardless of the annotation type.

\subsection*{Comparison with the proposed approach}

\begin{figure*}
\centering
\includegraphics[width=\textwidth]{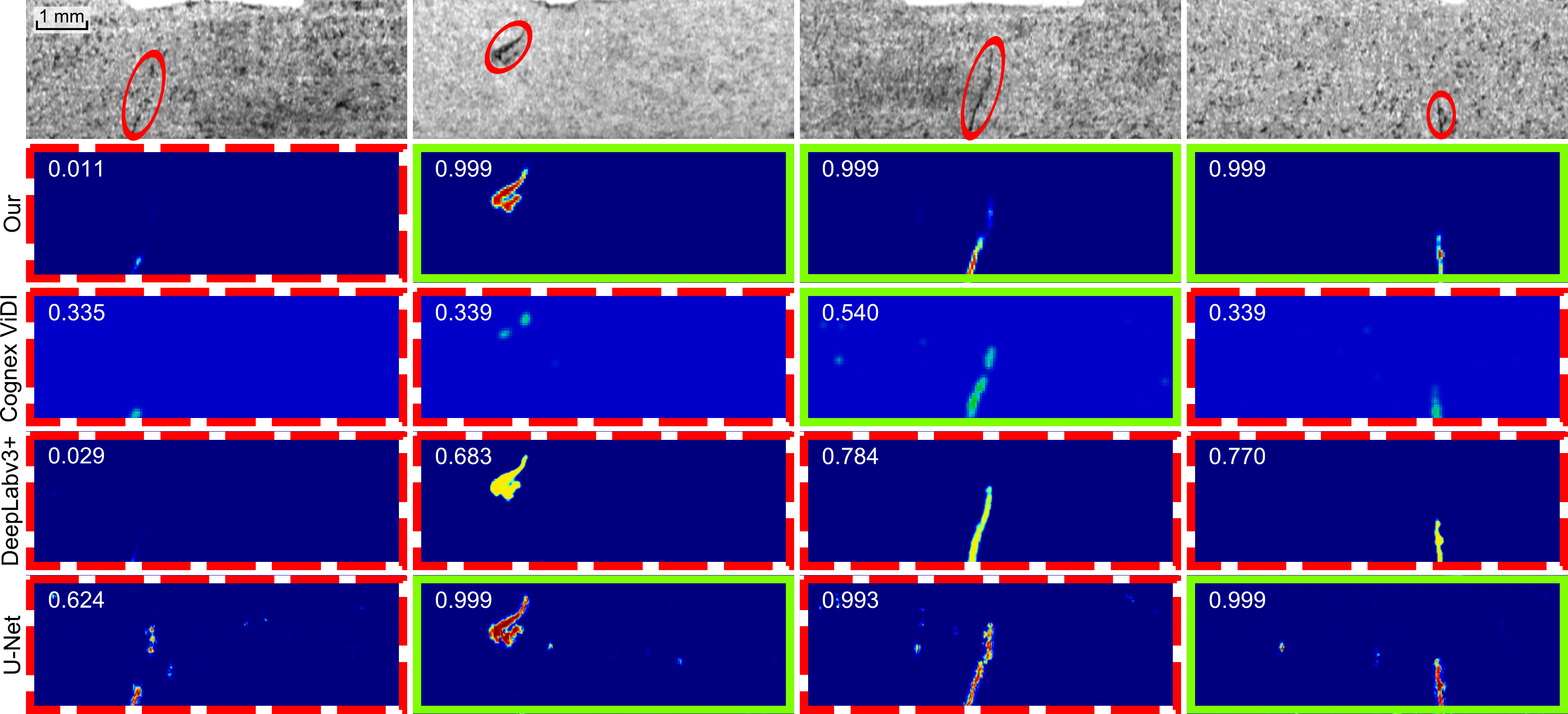}
\caption{Examples of true-positive (green solid border) and false-negative (red dashed border) detections with the segmentation output and the corresponding classification (the actual defect is circled in the first row) \label{fig:examples-fn}}
\end{figure*}

Finally, all three state-of-the-art approaches are compared against the network proposed in this paper. The state-of-the-art methods are compared with the combined segmentation and decision network. For a fair comparison all the methods reported in this section were selected based on the best-performing training setup from the evaluation in the previous sections. For all methods this included using the original image size ($1408\times512$ resolution), no input image rotation, using the smallest feature size of 20 pixels for the commercial software and using the cross-entropy loss function for all the remaining methods. For the annotation type different methods performed the best at different annotations. The commercial software and the proposed approach with the segmentation and decision network both achieved the best performance when trained on the \textit{dilate=5} annotations, while DeepLabv3+ and U-Net achieved the best results when trained using \textit{dilate=9} annotation. Selected configuration setups for each are shown in Table~\ref{tab:hyper-params}.

\paragraph{Results}

The results are presented in Fig.~\ref{fig:sota-results}. The proposed approach, shown in the left-most bar, outperformed all the state-of-the-art methods in all metrics. The commercial product performed the second best, while both standard segmentation methods preformed the worst, with the DeepLabv3+ architecture performing slightly better than the U-Net. Observing the number of miss-classifications at the ideal F-measure reveals that the proposed segmentation and decision network was able to reduce the miss-classification to only one false negative, while all the remaining methods introduced 5 or more miss-classifications. 

\begin{figure*}[!b]
\centering
\includegraphics[width=\textwidth]{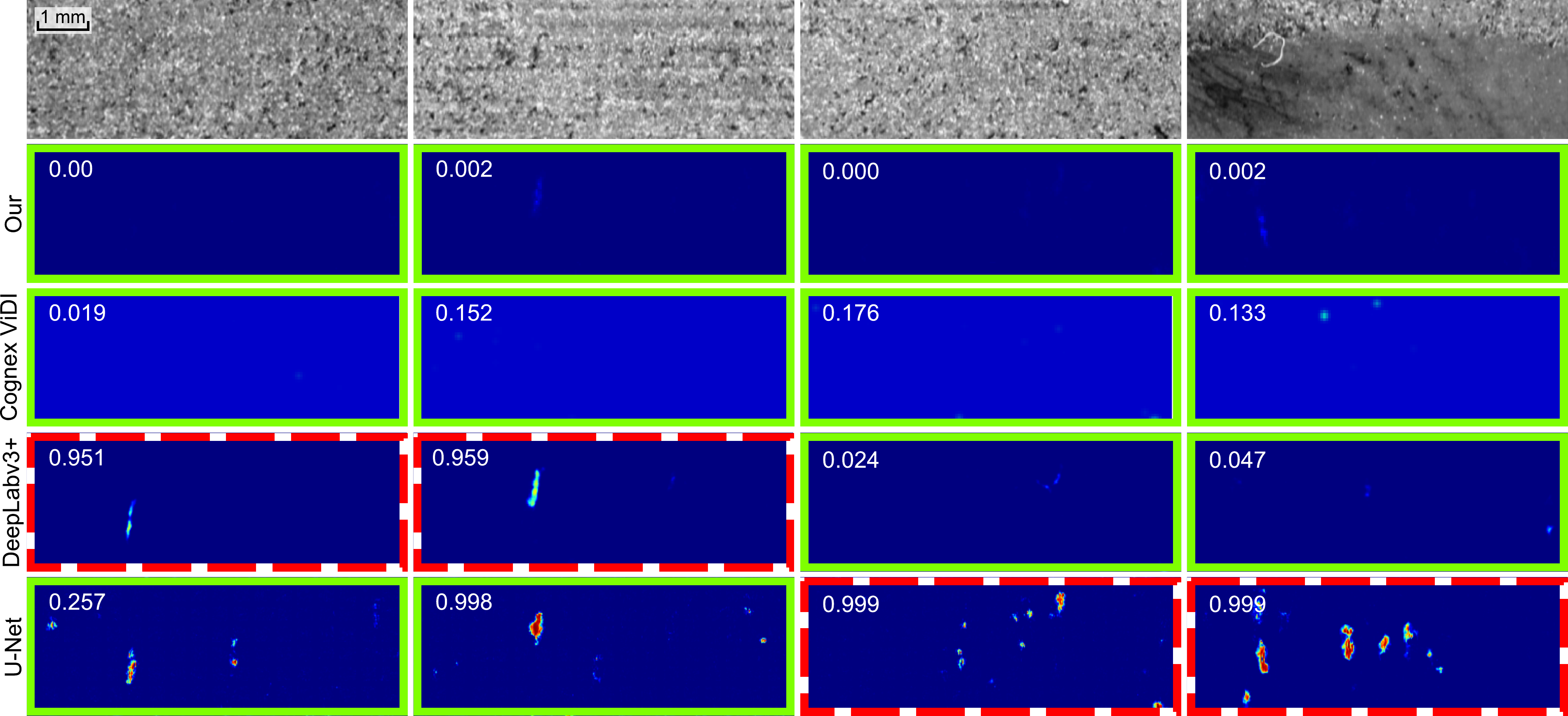}
\caption{Examples of true-negative (green solid border) and false-positive (red dashed border) detections with the segmentation output and the corresponding classification score\label{fig:examples-fp}}
\end{figure*}

Several miss-classified images for all methods are presented in Fig.~\ref{fig:examples-fn} and~\ref{fig:examples-fp}. True-positive and false-negative detections, as shown in Fig.~\ref{fig:examples-fn}, reveal a single missing detection for the proposed method in the first column. This sample contains a small defect that is difficult to detect and was not detected with any of the remaining methods as well. For the remaining examples the method proposed in this paper was able to correctly predict the presence of the defect, including a small defect seen in the last column. The proposed method was also able to localize the defects with excellent accuracy. Good localization can also be observed in the related methods; however, the prediction of the presence or absence of a defect was poor. Note that in some cases the score was large; however, to correctly separate all the defects from the non-defects the threshold needed to be set high as well, pointing to many false positives on the images without defects. This is well demonstrated in Fig.~\ref{fig:examples-fp}, showing several false detections. False positives with high scores can be observed in all the related methods, except in the method proposed in this paper and in the commercial software. In particular, U-Net returned an output with a large amount of noise, which prevented clean separation of true defects from false detections, even with the additional decision network. The proposed method, on the other hand, did not have any problems with the false positives and was correctly able to predict the absence of the defect in those images.

\begin{figure*}
\centering
\includegraphics[width=1\linewidth]{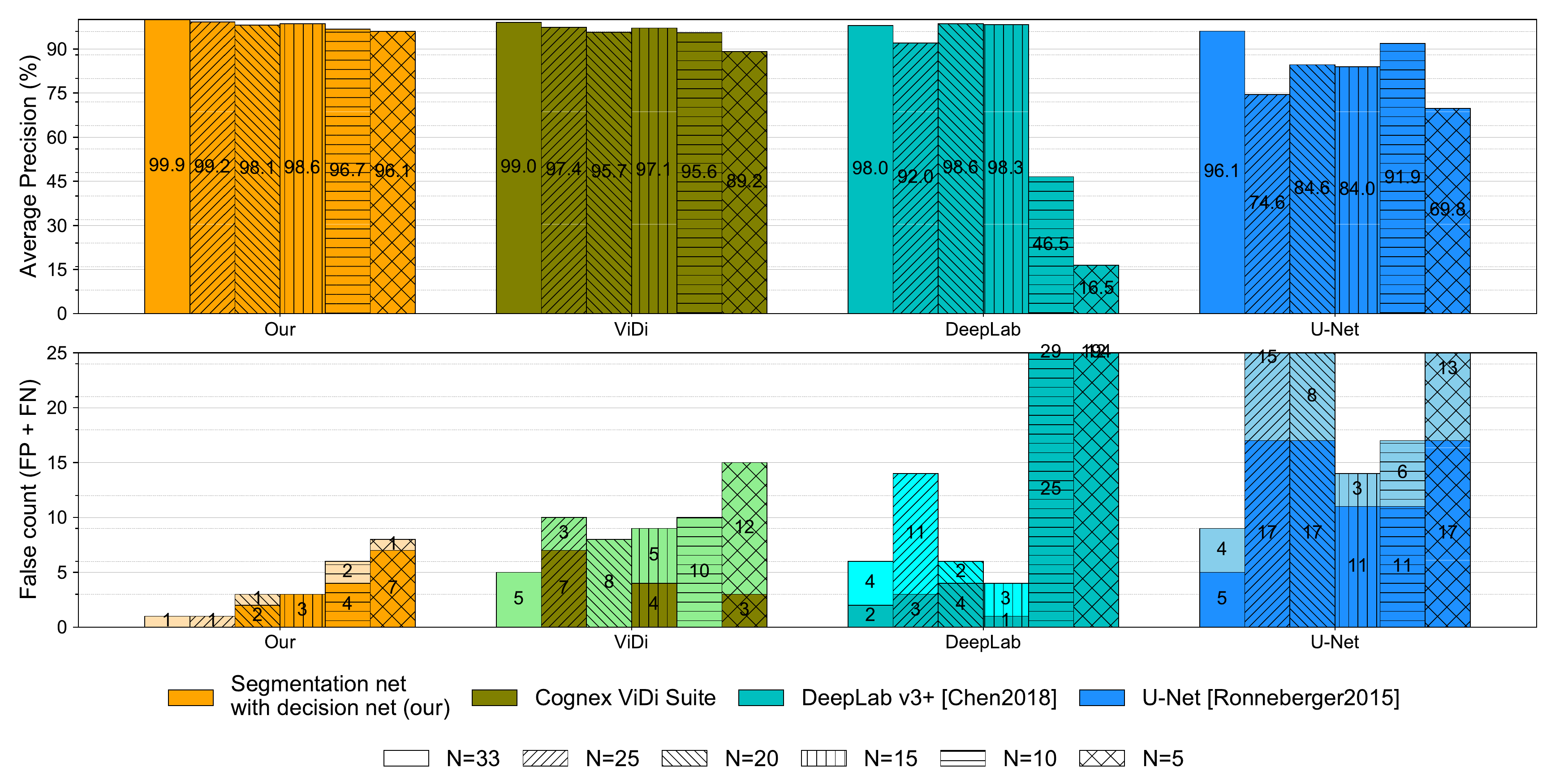}
\caption{Classification performance on KolektorSDD at varying number of positive (defective) training samples \label{fig:sota-num-training}}

\end{figure*}

\paragraph{Results in the context of an industrial environment}

When considering to use the proposed model in industrial settings it is important to ensure the detection of all the defected items, even at the expense of more false positives. Since previous metrics do not capture the performance under those conditions, this section sheds additional light on the number of false positives that would be obtained if a zero-miss rate would be required, i.e., if a recall rate of \num{100}{\%} is required. These false positives then represent the number of items that would be needed to be manually verified by a skilled worker and directly point to the amount of work required to achieve the desired accuracy. 

The results, as reported in the right-most graphs in Fig.~\ref{fig:sota-results}, show that the model as proposed in this paper introduces only 3 false positives at a zero-miss rate out of all 399 images. This represents \num{0.75}{\%} of all images. On the other hand, the related methods achieved worse results, with the commercial product requiring the manual verification of 7 images, while the standard segmentation networks required 68 and 108 manual verifications, for DeepLabv3+ and U-Net, respectively. Note that the results reported for both standard segmentations included using the proposed decision network. Using logistic regression instead of the proposed decision network resulted in significantly worse performance.

\subsection*{Sensitivity to the number of training samples}

In industrial settings a very important factor is also the required number of defective training samples, therefore we also evaluated the effect of smaller training sample size. The evaluation was performed using a 3-fold cross-validation with the same train/test split as used in all previous experiments, thus effectively using 33 positive (defective) samples in each fold when trained on all the training samples. The number of positive training samples was then reduced to effectively obtain the training size $N$ of 25, 20, 15, 10 and 5 samples for each fold, while the test set for each fold remained unchanged. 
The removed training samples were randomly selected, but the same samples were removed for all methods. The same training and testing procedure was followed as in all previous experiments. 

The proposed segmentation and decision network is compared with the commercial software Cognex ViDi Suite and two state-of-the-art segmentation networks. All methods are evaluated using the best performing training setup de\-ter\-mined in the experiments presented in the previous sections, i.e., using \textit{dilated=5} annotations (or dilated \textit{dilated=9} for segmentation networks), full image resolution, cross-entropy loss and no image rotation. Results are reported in Figure~\ref{fig:sota-num-training}. The proposed segmentation and decision network retains the same result of over 99\% AP and a single miss-classification when using only 25 defective training samples. When using even less training samples the results drop, but the proposed method still achieves AP of around 96\% when only 5 defective training samples were used. More pronounced drop in performance can be observed for the Cognex ViDi Suite, however, in this case the results drop already at N=25 to AP of 97.4\%. When using only 5 defective training samples the commercial software achieved AP of slightly below 90\%. The same trend is observed in the number of miss-classifications depicted in the bottom half of Figure~\ref{fig:sota-num-training} with the dark colors representing false positives and the light colors representing false negatives.

The DeepLab v3+ and U-Net, on the other hand, perform worse than the proposed approach when less training samples are used. The performance of U-Net quickly drops, while Deep\-Lab retains fairly good results even for only 15 defected training samples. Note that the performance at 20 and 15 defected training samples slightly outperforms the results obtained with all training samples, indicating that Deep\-Lab is fairly sensitive to specific training examples and removing such samples helps in improving the performance. U-Net is significantly more sensitive to the decrease of the number of training samples; the results varied from 75\% to slightly above 90\% in average precision. However, for 10 and 5 defective training samples, DeepLab performed the worst with AP of only 46\% and 16\%, respectively.

Overall, the experimental results show that the proposed method retains superior and stable performance also when smaller number of training samples are available.

\subsection*{Computational cost}

\begin{figure}
\centering
\includegraphics[width=0.95\columnwidth]{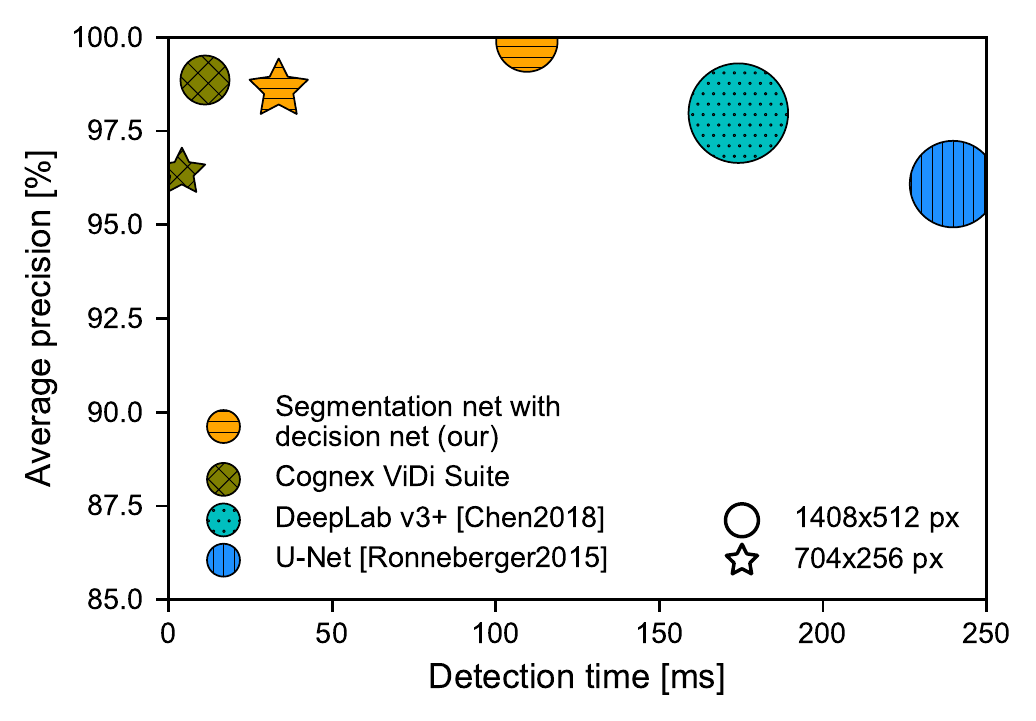}
\caption{Detection (forward pass) time with respect to the classification performance for a single image \label{fig:sota-speed}}
\end{figure}

The approach proposed in this paper is superior to the state-of-the-art segmentation methods in terms of computational cost and is competitive with the commercial software. For\-ward-pass times with respect to the average precision are reported in Fig.~\ref{fig:sota-speed}. Results were obtained on a single NVIDIA TITAN X (Pascal) GPU. The proposed method is shown to be significantly faster than DeepLab v3+ and U-Net, with a better accuracy as well. This is achieved with a smaller number of parameters, which is reflected in the marker size in Fig.~\ref{fig:sota-speed} and is shown in Table~\ref{tab:sota-eval} as well. This performance is achieved using only 15.7 mio parameters for the proposed model, while U-Net and DeepLab v3+ have more than twice as many parameters, with 31.1 mio and 41.1 mio parameters, respectively. The number of parameters for the Cognex ViDi Suite is not publicly available. 
The proposed method and commercial software are also shown at half the resolution depicted with the \textit{star} marker in Fig.~\ref{fig:sota-speed}. This shows that the proposed method results in a 3-times faster forward pass than with full resolution---33 ms for the half-resolution and 110 ms for the full-resolution image. The fastest performance is achieved with the commercial software, Cognex ViDi Suite, with 10 ms per image. However, when using half the image resolution the proposed best-performing model achieves a similar performance with only a slightly larger computational cost. Note that the proposed model achieved this performance in the TensorFlow framework without applying any computational optimization, while it can be safely assumed that the commercial software has been highly optimized to reduce the computational cost as much as possible.

\begin{table}
\caption{Comparison with the state-of-the-art methods in the number of learnable parameters and average precision\label{tab:sota-eval}}
\centering
\begin{tabularx}{\columnwidth}{Xccc}
\toprule
\textbf{\textit{Method}} & \makecell[l]{Number of\\parameters} & \makecell[l]{Average\\precision}  \\
\midrule
Segmentation/decision net (our)  		& \textbf{15.7 mio} & \textbf{99.9}\\
Cognex ViDi Suite 					& N/A &  98.9 \\
DeepLab v3+~\citep{Chen2018}			& 41.1 mio & 97.9 \\
U-Net~\citep{Ronneberger2015}		& 31.1 mio & 96.1 \\

\bottomrule
\end{tabularx}
\end{table}

\begin{figure*}
\centering
\includegraphics[width=\textwidth]{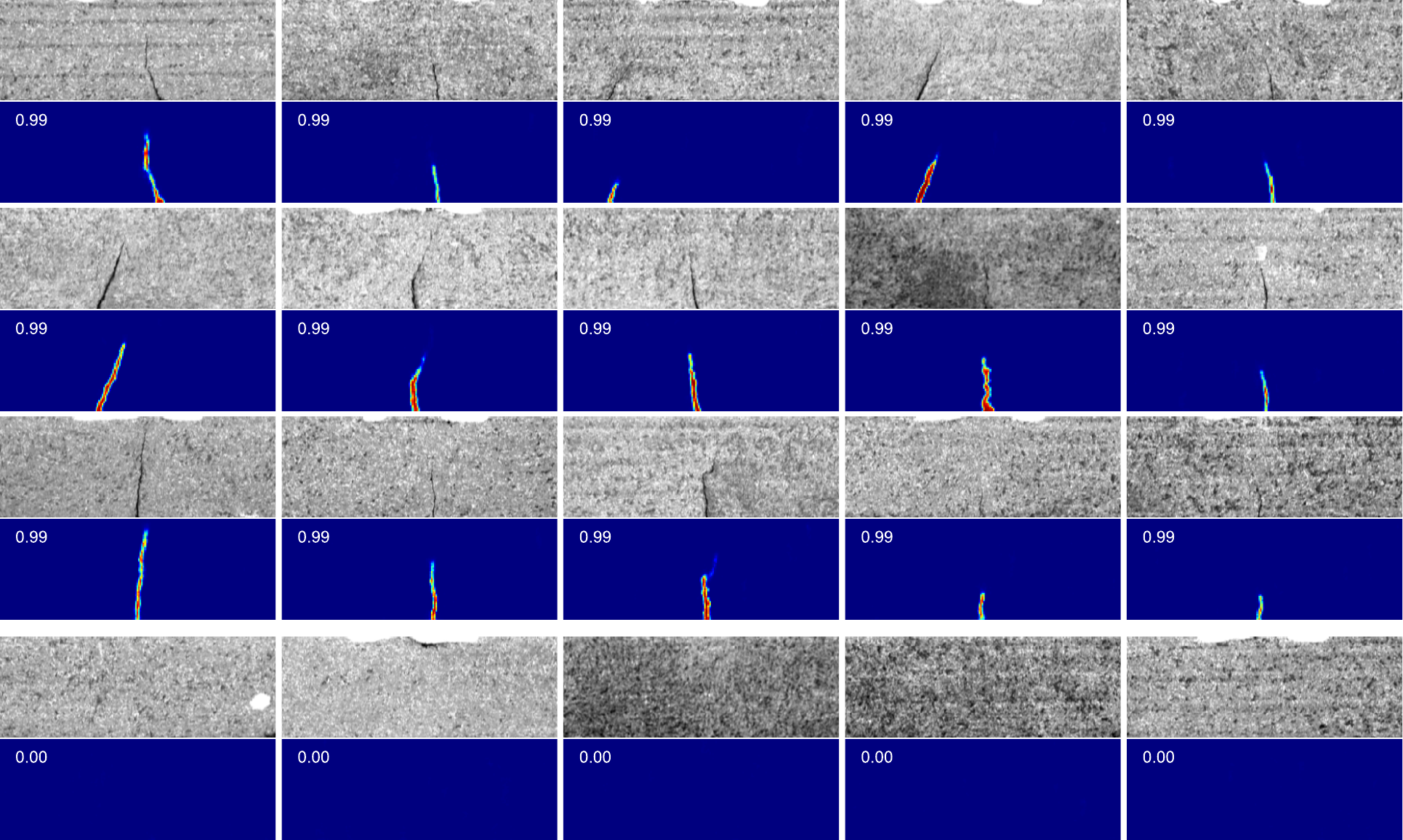}
\caption{Examples of true-positive (the upper three rows) and true-negative (the bottom row) detections on KolektorSDD with the proposed approach (classification score is depicted in the top-left corner for each example) \label{fig:examples-tp-our}}
\end{figure*}

\section*{Discussion and conclusion}\label{sec:conclusion}

This paper explored a deep-learning approach to surface-defect detection with a segmentation network from the point of view of specific industrial application. A two-stage approach was presented. The first stage included a segmentation network trained on pixel-wise labels of the defect, while the second stage included an additional decision network build on top of the segmentation network to predict the presence of the anomaly for the whole image. An extensive evaluation of the proposed approach was made on a semi-finished industrial product, i.e., an electrical commutator, where the surface defects appeared as fractures of the material. This problem domain has been made publicly available as a benchmark dataset, termed the Kolektor Surface-Defect Dataset (KolektorSDD). The proposed approach was compared on this domain with several state-of-the-art methods, including proprietary software and two standard segmentation methods based on deep learning. 

The experiments on KolektorSDD demonstrated that the proposed model achieves significantly better results than related methods with only one miss-classification, while the related methods achieve five or more miss-cla\-ssi\-fi\-cations. This can be attributed to the proposed two-stage design with the segmentation and the decision network, as well as to the improved receptive field size and an increased capacity to capture the fine details of the defect. The related methods are missing some of those characteristics. For instance, the worst-performing segmentation method, U-Net, has a limited receptive field size, with only 45 pixels versus 205 pixels of the proposed method. Although DeepLabv3+ improves the receptive field size, it does this at the expense of too many parameters, which cause the model to overfit, despite being pre-trained on separate datasets. 

On the other hand, it is difficult to assess the differences with the commercial software, since the details of the method are not publicly known. Nevertheless, the experiments show that the commercial software performs significantly worse than the proposed method when using lower-resolution images. This experiment is an indication that the commercial software struggles to capture finer details of the defect and requires a higher resolution for good performance. However, it still cannot attain the same performance as the proposed method even when high-resolution images are used. 

The performance of the proposed method was achieved by learning from only 33 defective samples. Several examples of correct classification are depicted in Figure~\ref{fig:examples-tp-our}. Moreover, using only 25 defective samples showed that good performance can still be attained, while related methods achieved worse results in this case. This indicates that the proposed deep-learning approach is suitable for the studied industrial application with a limited number of defected samples available. Moreover, to further consider applications for the industrial environment, three important characteristics were evaluated: (a) the performance to achieve 100\% detection rate, (b) details of the annotation and (c) the computational cost. In terms of the performance to achieve a 100\% detection rate the proposed model has been shown to require only three images for the manual inspection out of all 399 images, leading to a 0.75\% inspection rate. Large and coarse annotations also turned out to be sufficient to achieve a performance similar to the one with finer annotations. In some cases larger annotations even resulted in better performance than using fine annotations. This conclusion is seemingly counter-intuitive; however, a possible explanation can be found in the receptive field size used to classify each pixel. The receptive field for a pixel that is slightly away from the defective area will still cover part of the defective area and can therefore contribute towards finding features that are important for their detection, if they are annotated correctly. This conclusion can result in reduced manual work when adapting methods to new domains and will lead to reduced labor costs and the increased flexibility of production lines. 

The proposed approach is, however, limited to the specific type of tasks. In particular, the architecture was designed for tasks that can be framed as a segmentation problem with pixel-wise annotation. Other quality-control problems exist for which a segmentation-based solution is less suitable. For instance, quality control of complex 3D objects may require detection of broken or missing parts. Such problems could be addressed by detection methods, such as Mask R-CNN~\citep{He2017}.

This study demonstrated the performance of the proposed approach on a specific task (crack detection) and on a specific surface type, but the architecture of the network was not designed for this specific domain only. Learning on new domains is possible without any modification. The architecture can be applied to images that contain multiple complex surfaces, or it can be applied to detect other different defect patterns, such as scratches, smudges or other irregularities, providing that a sufficient number of defected training samples is available and that the task of the particular defect detection can be framed as a surface segmentation problem. However, to further evaluate this, new datasets are needed. To the best to our knowledge, the DAGM dataset~\citep{Weimer2016} is the only publicly available annotated dataset with a diverse set of surfaces and defect types suitable for evaluation of learning-based approaches. The proposed approach achieves perfect results on this dataset, which is, however, synthetically generated and also saturated according to the obtained results. Future effort should, therefore, be focused also on acquiring new complex datasets based on real-world visual inspection problems, where deep-learning (and other) methods could be realistically evaluated in full extent; the dataset presented in this paper is a first step in this direction.


\begin{acknowledgements}
This work was supported in part by the following
research projects and programs: GOSTOP program C3330-16-529000 co-financed by the Republic of Slovenia and the European Regional Development Fund, ARRS research project J2-9433 (DIVID), and ARRS research programme P2-0214. We would also like to thank the company Kolektor Orodjarna d. o. o. for providing images for the proposed dataset as well as for providing high quality annotations. 
\end{acknowledgements}

{\small
\bibliographystyle{spbasic}
\bibliography{library}
}

\end{document}